\documentclass{article}
\usepackage[table]{xcolor}
\usepackage{booktabs}
\usepackage{graphicx}
\usepackage{booktabs}
\usepackage{float}
\usepackage{placeins}
\usepackage{graphicx}
\usepackage{amssymb}
\usepackage{array}
\usepackage{longtable}
\usepackage{capt-of}
\usepackage[final]{corl_2026}
\usepackage{xurl}
\usepackage{xspace}
\usepackage{pifont}
\newcommand{\cmark}{\ding{51}}
\newcommand{\xmark}{\ding{55}}
\newcommand{\PAPERNAME}{HARP-Bench\xspace}
\definecolor{posgreen}{RGB}{35,120,70}
\definecolor{negred}{RGB}{160,60,60}
\hypersetup{
  pdftitle={HARP-Bench: Benchmarking Robot Manipulation Around Active Humans},
  pdfauthor={Qinhong Zhou, Riley Zong, Chunru Lin, Diyi Hu, Shivraj Singh Bhatti, Hongxin Zhang, Lin Qiu, Hanqing Zeng, Chuang Gan}
}

\title{\PAPERNAME: Benchmarking Robot Manipulation Around Active Humans}

\author{
\normalfont
Qinhong Zhou$^{1,\dagger}$ \quad
Riley Zong$^{2,\dagger}$ \quad
Chunru Lin$^{1}$
\\[0.15em]
Diyi Hu$^{4}$ \quad
Shivraj Singh Bhatti$^{1}$ \quad
Hongxin Zhang$^{1}$
\\[0.15em]
Lin Qiu$^{3}$ \quad
Hanqing Zeng$^{3}$ \quad
Chuang Gan$^{1}$
\\[0.35em]
\small
$^{1}$University of Massachusetts Amherst \quad
$^{2}$University of Minnesota Twin Cities
\\
\small
$^{3}$Meta \quad $^{4}$Independent Researcher
}

\begin{document}
\maketitle
\begingroup
\renewcommand{\thefootnote}{\fnsymbol{footnote}}
\footnotetext[2]{Equal contribution.}
\endgroup

\begin{abstract}
Recent robot manipulation benchmarks have enabled systematic evaluation of general-purpose robot policies across diverse objects, scenes, instructions, and embodiments. However, most existing benchmarks still assume that the robot is the only active agent and that task semantics are determined primarily by the instruction and object state. This assumption overlooks a common source of real-world variation: humans actively sharing the workspace. In such settings, humans may reveal an implicit goal, complete part of the task, temporarily change object availability, or constrain safe robot motion while working nearby.
We introduce \emph{HARP-Bench}, a benchmark for evaluating robot manipulation policies in workspaces shared with active humans. HARP-Bench treats human activity as part of the manipulation problem, rather than as background scene variation. It organizes human-present manipulation into four conditions: \emph{Cueing}, \emph{Contribution}, \emph{Interruption}, and \emph{Co-presence}. These conditions are instantiated through everyday manipulation tasks, including paired human-present variants for controlled comparison with standard manipulation. HARP-Bench also provides a demonstration-generation pipeline for dynamic human-present settings.
We evaluate VLA models, imitation-learning methods, and heuristic policies on \PAPERNAME. Human-present variations challenge current policies, and higher task success does not consistently improve safety. Quantitative failure analysis identifies manipulation and recovery as the dominant remaining challenges. Training with \PAPERNAME demonstrations improves human-present performance in simulation. A real-world study on one task family shows improved performance under human contribution, but not interruption, providing initial evidence of the data's utility and identifying limits of transfer.
Code is available at \url{https://github.com/UMass-Embodied-AGI/harp}.
\end{abstract}

\keywords{human-robot interaction, dataset, benchmark}
\section{Introduction}

Recent robot manipulation benchmarks~\cite{rlbench,robosuite,calvin,libero,maniskill3,robocasa,vlabench, robowits} have become increasingly scalable and systematic, evaluating policies across diverse tasks, objects, scenes, instructions, embodiments, and temporal horizons. Despite this progress, most benchmarks vary the passive conditions under which a robot acts~\cite{robotwin2, roboeval}. Objects, layouts, and instructions may change, but the robot is typically the only agent that changes the workspace. Consequently, the meaning of a task is largely fixed by the language instruction and the current object state. This assumption breaks down in workspaces shared with people, where human actions can alter what the robot should do, which objects are available, and where the robot can move safely.

In human-present workspaces, the robot's next action may depend on human activity rather than on object state alone. For example, a person reaching toward a task-relevant object may require the robot to pause, while a person completing part of the task may require the robot to update its plan. More broadly, human activity can change the robot's goal, task progress, object availability, and motion constraints. Evaluating such behaviors requires benchmarks that go beyond standard object-centric manipulation.

Several recent benchmarks have begun to consider humans in embodied evaluation. However, most of these settings either operate at a high level of abstraction~\cite{partnr, humanthor}, where interaction is represented through planning or skill-level actions, or remain limited in scale and visual richness for evaluating modern vision-language-action manipulation policies~\cite{humanrobotgym}. This leaves a gap for scalable, visually grounded benchmarks that test autonomous manipulation policies in human-present workspaces.

We introduce \PAPERNAME, a simulation benchmark for evaluating autonomous robot manipulation policies in human-present workspaces. \PAPERNAME organizes human activity into four evaluation conditions: \emph{Cueing}, \emph{Contribution}, \emph{Interruption}, and \emph{Co-presence}. These conditions capture whether human behavior indicates the robot's objective, completes part of the task, changes object availability or execution timing, or adds motion constraints while working nearby. \PAPERNAME implements these settings with animated human avatars and closed-loop avatar events during policy rollouts, together with RGB-D observations, natural-language instructions, safety-aware metrics, and data interfaces for VLA and imitation-learning policies.

Beyond task design, \PAPERNAME provides an evaluation and data-generation pipeline for human-present manipulation. Avatar behaviors are triggered during closed-loop policy rollouts, so policies are evaluated against human activity at execution time. \PAPERNAME also supports scripted demonstration collection, VLA-oriented trajectory recording, and a LeRobot-compatible data interface, with additional conversion paths for other VLA and imitation-learning pipelines.

We use \PAPERNAME to evaluate two expert references and representative imitation-learning and VLA policies under the same human-present evaluation protocol. We report task success across the four conditions, together with aggregate safety metrics and a quantitative failure breakdown. Higher success does not consistently improve safety, and manipulation/recovery dominates the failures of both imitation-learning and VLA policies. Training on \PAPERNAME demonstrations improves performance on matching human-present variants, suggesting that the benchmark can support both evaluation and data generation for robot policies around active humans.

Our contributions are as follows:

\noindent \textbf{Benchmark.}
We introduce \PAPERNAME, a simulation benchmark with 37 tasks for evaluating robot manipulation in human-present workspaces, organized into four conditions: \emph{Cueing}, \emph{Contribution}, \emph{Interruption}, and \emph{Co-presence}.

\noindent \textbf{Evaluation.}
\PAPERNAME provides closed-loop avatar-driven evaluation, where human activity occurs during policy rollouts. Policies are evaluated by both task success and safety metrics.

\noindent \textbf{Data generation.}
\PAPERNAME supports scripted demonstration collection, VLA-oriented trajectory recording, and a LeRobot-compatible data interface, with additional conversion paths for other VLA and imitation-learning pipelines.
\vspace{-2mm}
\section{Related Works}
\vspace{-2mm}

\subsection{Robot Manipulation Foundation Models, Data, and Benchmarks}

Recent robot learning has moved toward general-purpose manipulation policies that map visual observations and language instructions to robot actions. This includes imitation-learning methods, diffusion-based policies, and vision-language-action models trained across diverse tasks and embodiments~\cite{act,diffusionpolicy,dp3,rt2,octo,openvla,pi0,pi05,rdt, lee2025molmoact, bjorck2025gr00t}. Their progress has been closely tied to the growth of large-scale robot datasets and open training infrastructure, which provide diverse experience across environments, objects, tasks, and robot embodiments~\cite{bridgedatav2,droid,openxembodiment,lerobot}. As these policies become more general, benchmarks are increasingly needed to evaluate not only whether a robot can solve individual tasks, but also how policies generalize across task structure, language, visual scenes, embodiments, and temporal horizons.

Existing manipulation benchmarks have expanded along these axes, from reusable simulated environments for language-conditioned manipulation to large-scale benchmarks with diverse tasks, objects, scenes, visual realism, long-horizon structure, and data generation~\cite{rlbench,robosuite,calvin,libero,liberox,maniskill3,robocasa,robocasa365,robotwin2,vlabench,vlaarena, atreya2025roboarena}. However, most of these benchmarks still define variation over passive task factors: objects, layouts, instructions, and embodiments may change, while the robot remains the only active agent that changes the workspace. \PAPERNAME builds on this line of benchmark and data infrastructure, but introduces human-present workspaces as an additional evaluation setting, where human activity can affect the robot's objective, object availability, task progress, timing, and safe motion.

\subsection{Human-Centered Embodied Benchmarks and Avatar Simulation}

Recent work has introduced humans into embodied and robotic benchmarks through household goals, collaborative planning, human-in-the-loop interaction, high-level assistance, and responsible decision making~\cite{behavior1k,partnr,humanthor,m4bench,niabench,responsiblerobotbench, zhang2024building, zhang2025combo, carroll2019utility}. However, these settings often abstract away low-level manipulation execution, focusing instead on planning, user interaction, assistance policies, or risk-aware reasoning and detection. As a result, they do not directly evaluate autonomous manipulation policies acting around active humans from visual observations and language instructions.

Low-level HRC work is more directly related, but existing simulation benchmarks are limited in task scale and visual richness~\cite{humanrobotgym}, while real-world studies are typically task-specific and difficult to scale~\cite{mosaic, pace,decaf}. This leaves a gap for scalable, visually grounded simulation benchmarks for human-present manipulation. Avatar simulation and motion resources make it possible to generate repeatable human activity in simulation~\cite{virtualcommunity,amass,mixamo}. \PAPERNAME builds on these tools to construct controlled human-present manipulation conditions with closed-loop avatar events, safety-aware metrics, and data-generation interfaces for imitation-learning and VLA pipelines.
\vspace{-2mm}
\section{HARP-Bench: Robot Manipulation around Active Humans}
\vspace{-2mm}

\begin{figure}[t]
    \centering
    \includegraphics[width=0.8\linewidth]{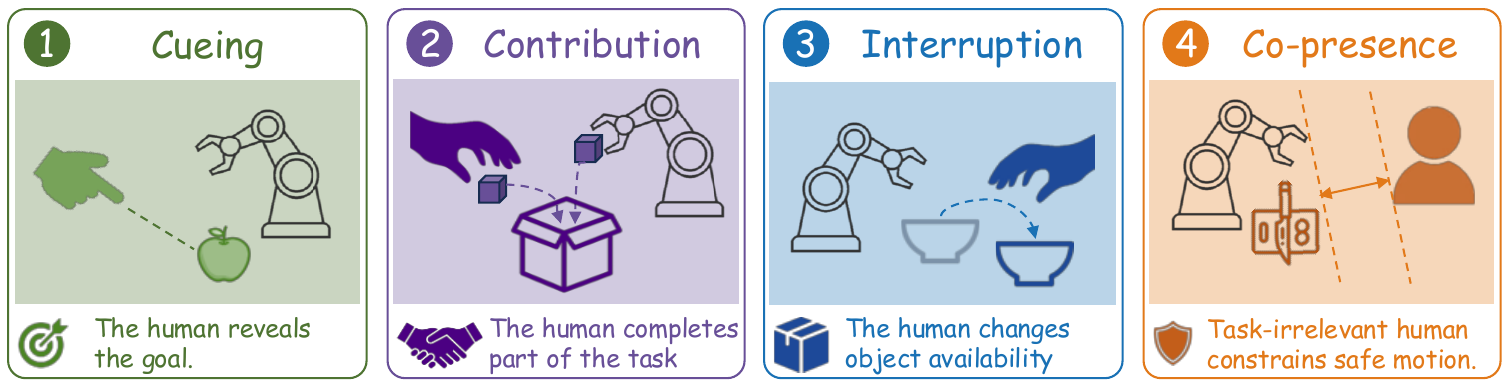}
    \caption{
    \textbf{Task design in \PAPERNAME.} Human activity is organized into four evaluation conditions. 
    Cueing changes what the robot should do, Contribution changes which subtasks remain, Interruption changes when or how the robot can act, and Co-presence adds motion constraints while leaving the original goal unchanged.
    }
    \label{fig:task_taxonomy}
    \vspace{-4mm}
\end{figure}

\PAPERNAME is a simulation benchmark for evaluating robot manipulation policies in human-present workspaces. It extends standard object-centric manipulation tasks with avatar-driven human activity that can change the robot's objective, object availability, action timing, and motion constraints. The benchmark consists of 37 benchmark tasks organized around four human-present conditions, and provides closed-loop evaluation, safety-aware metrics, and data-generation interfaces for imitation-learning and VLA pipelines. Table~\ref{tab:scope_comparison} summarizes the scope of \PAPERNAME relative to related benchmarks.

\subsection{Task Design}

\PAPERNAME is designed around the ways human activity changes a manipulation task. In standard object-centric benchmarks, the robot's objective is usually specified by the language instruction and the object configuration. In human-present workspaces, the same object-level skill may require different behavior depending on what the human is doing. We organize these differences into four evaluation conditions: \emph{Cueing}, \emph{Contribution}, \emph{Interruption}, and \emph{Co-presence}. Figure~\ref{fig:task_taxonomy} illustrates these conditions with representative task instances.

\emph{Cueing} tasks evaluate whether the robot can infer the appropriate objective from human activity. For example, the human may carry an object, offer an object, or perform an ongoing activity that indicates what the robot should do next. \emph{Contribution} tasks evaluate whether the robot can account for human-completed subtasks, such as a human placing one item while the robot completes the remaining items. \emph{Interruption} tasks evaluate whether the robot can adapt when the human temporarily changes execution conditions, such as inspecting, holding, or occupying a task-relevant object. \emph{Co-presence} tasks evaluate whether the robot can complete its original manipulation goal while a human performs unrelated work nearby.

These conditions connect to established HRI problems: human-intent inference and human-action prediction for Cueing~\cite{hoffman2024inferring}; human-progress estimation and timely proactive assistance for Contribution~\cite{pace}; reactive task replanning and recovery under asynchronous human actions for Interruption~\cite{decaf}; and human-aware motion planning and shared-workspace safety for Co-presence~\cite{lasota2015analyzing}. They respectively capture changes to the goal, remaining subtasks, action availability or timing, and motion constraints. This organization isolates effects on manipulation rather than constituting an exhaustive taxonomy of human behavior.

The 37 tasks are task-condition instances, with multiple human-present variants sharing an underlying manipulation skill. This paired design supports comparisons while keeping object-level goals comparable; it does not imply 37 distinct manipulation skills. The suite includes tabletop object manipulation and articulated-object interactions such as microwave and cabinet operation.

\begin{table}[t]
    \centering
    \small
    \setlength{\tabcolsep}{4pt}
    \caption{
    \textbf{Comparison with related benchmarks.} Manip. denotes low-level manipulation; Human denotes active humans in the workspace; Active denotes human activity during policy rollout; Visual denotes VLA-style visual observations; Data denotes demonstration or dataset generation support.
    }
    \label{tab:scope_comparison}
    \begin{tabular}{lccccc}
        \toprule
        Benchmark & Manip. & Human & Active & Visual & Data \\
        \midrule
        RoboTwin2~\cite{robotwin2}  & \cmark & \xmark         & \xmark         & \cmark & \cmark \\
        VLABench~\cite{vlabench}   & \cmark & \xmark         & \xmark         & \cmark & \cmark \\
        RoboCasa~\cite{robocasa}   & \cmark & \xmark         & \xmark         & \cmark & \cmark \\
        RoboWits~\cite{robowits} & \cmark & \xmark & \xmark & \cmark & \cmark \\
        PARTNR~\cite{partnr}     & \xmark         & \cmark & \cmark         & \xmark         & \xmark         \\
        HumanTHOR~\cite{humanthor}  & \xmark     & \cmark & \cmark & \cmark     & \xmark         \\
        HR-Gym~\cite{humanrobotgym}     & \cmark & \cmark & \xmark     & \xmark         & \xmark         \\
        \midrule
        \PAPERNAME         & \cmark & \cmark & \cmark & \cmark & \cmark \\
        \bottomrule
    \end{tabular}
    \vspace{-2mm}
\end{table}

\vspace{-2mm}
\subsection{Closed-Loop Evaluation with Simulated Human Activity}
\label{sec:bencheval}

\begin{figure}[t]
    \centering
    \includegraphics[width=\linewidth]{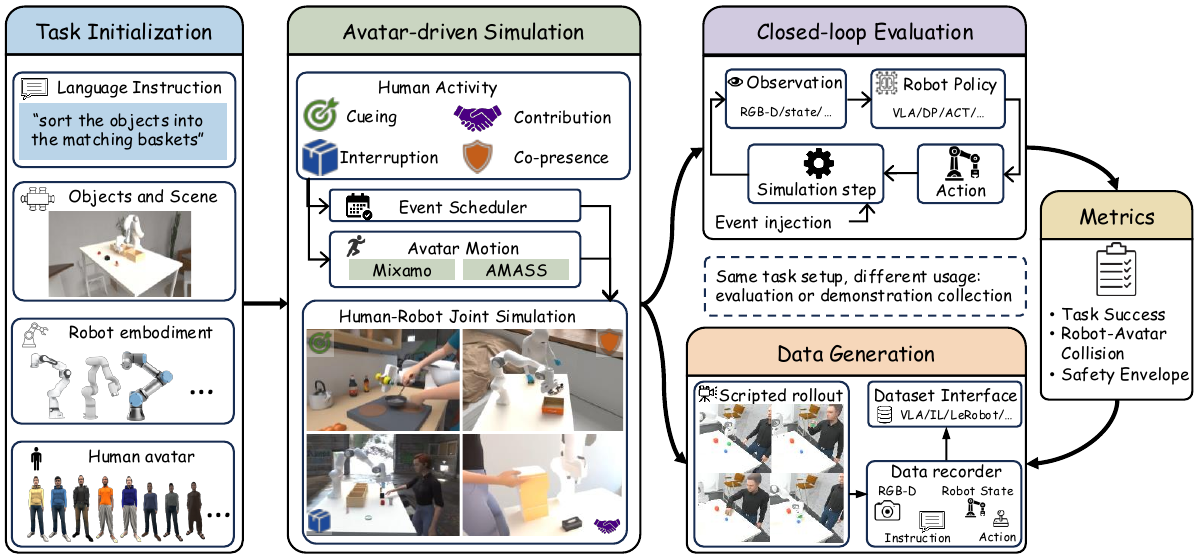}
    \caption{
    \textbf{Overview of the \PAPERNAME pipeline.} Each task initializes a scene with objects, a robot embodiment, and avatar-driven human activity. The same task setup supports two usage modes: closed-loop evaluation, where avatar events occur during execution and metrics are computed, and data generation, where scripted rollouts are recorded for robot learning.
    }
    \label{fig:framework}
    \vspace{-4mm}
\end{figure}

As Figure~\ref{fig:framework} shows, \PAPERNAME uses animated human avatars to model task-relevant human activity in shared manipulation workspaces. Our avatar simulation follows the avatar-robot joint simulation framework used in Virtual Community~\cite{virtualcommunity}, adapted here for tabletop robot manipulation. Each avatar is represented as a skinned human model driven by retargeted motion clips. Avatar appearances are drawn from an avatar pool. The motion library combines Mixamo's motion-captured animations performed by professional actors~\cite{mixamo} with AMASS-derived motions, which originate from optical motion-capture datasets unified in a common body-model representation~\cite{amass}. The avatar controller supports idle and walking motions, object inspection, pick-and-place motions, drawer-pull motions, and hand-object attachment for one- or two-handed interactions.

For each task, avatar behavior is specified through manipulation-relevant events. The avatar may approach the table, hold or offer an object, inspect a task object, place an item, open a drawer, or perform unrelated nearby work. For human-object interactions, objects can be attached to the avatar hand at specified animation frames and detached later. For some human-side scene changes, such as opening a drawer, the non-robot articulation is synchronized with the avatar motion to produce stable and repeatable scenarios. Robot-side manipulation remains physics-based, while human-side events are scripted to provide controlled evaluation conditions.

Avatar events occur during closed-loop robot execution. Cueing typically begins with a motion indicating the intended action. Contribution events begin at an early randomized time, Interruption events can depend on the gripper position near a relevant substep, and Co-presence uses randomly selected nearby motions. Although the source motions are based on human performances, task events and timing remain scripted for controlled, repeatable evaluation. This does not capture the full variability of real human behavior or continuous human-robot adaptation; adaptive human models remain future work.

\noindent \textbf{Evaluation metrics.}
Task success uses task-specific completion criteria, such as object placement, pose, height, or arrangement. An analytic collision checker computes signed distances between sampled robot-link geometry and query-only avatar body geometry; a robot-avatar collision makes the episode a failure. We additionally report three safety metrics: \emph{Collision}, the fraction of episodes with a robot-avatar collision; \emph{Min. Dist.}, the mean of each episode's minimum robot-avatar distance; and \emph{Violat.}, the fraction of timesteps with distance $d_t<\tau$, using a benchmark-specific proximity threshold $\tau=0.10$\,m. These metrics distinguish collision incidence, closest approach, and exposure to close proximity. The proximity criterion is inspired by the speed-and-separation monitoring principle in ISO/TS~15066:2016~\cite{iso15066}. The fixed 10\,cm threshold is our evaluation convention, not a standard-prescribed protective distance; it neither implements a speed-dependent protective separation calculation nor establishes real-world safety compliance.

\vspace{-2mm}
\subsection{Data Generation and Policy Interfaces}

As shown in Fig.~\ref{fig:framework}, the data-generation branch uses the same task initialization and avatar simulation, but records scripted rollouts as training trajectories. For data generation, each task provides scripted rollouts that produce successful demonstrations under the corresponding human-present condition. These rollouts are used to collect videos and trajectory data, but are separate from policy evaluation, where learned policies act in a closed loop with avatar events triggered during execution.

For VLA and imitation-learning pipelines, \PAPERNAME records trajectories with visual observations, robot state, language instructions, and action labels. The recorder stores synchronized camera observations and robot measurements at the policy frequency, so the resulting data can be used directly for policy training or offline analysis.

\PAPERNAME provides a LeRobot~\cite{lerobot}-compatible data interface, with additional conversion paths for other VLA and imitation-learning pipelines. It also exposes a common policy interface for evaluating learned policies from visual observations and language instructions, allowing different policy families to be compared under the same human-present task and safety metrics.
\vspace{-2mm}
\section{Experiments}
\vspace{-2mm}

\subsection{Experimental Setup}

\noindent \textbf{Data and splits.}
We train and evaluate policies using demonstrations collected from scripted rollouts in \PAPERNAME. The training set contains 11,100 demonstrations across 37 tasks, with 300 per task, including 7 Cueing, 10 Contribution, 10 Interruption, and 10 Co-presence tasks. We hold out 100 evaluation seeds per task and report results over the corresponding 100 evaluation episodes. Unless otherwise specified, learned policies are trained with the same visual observations, language instructions, and action representation. VLA instructions use human-authored templates; action labels come from the recorded robot trajectories.

\noindent \textbf{Evaluation protocol.}
All policies are evaluated with closed-loop avatar events under the same task initialization and success criteria. We report task and condition-level success, aggregate safety metrics as defined in Sec.~\ref{sec:bencheval}, and primary failure categories. Evaluation seeds vary rollout conditions; they do not measure variability across independently trained policies.

\noindent \textbf{Baselines.}
We compare two expert references with representative imitation-learning and VLA policies.

\noindent \textbf{Expert references.}
Both experts observe object poses and robot state, and share task-specific state machines, annotated grasps, inverse kinematics, and RRT motion planning. The \emph{full-state expert} additionally observes avatar pose, event start/end states, and the current event-conditioned subgoal. It serves as a privileged feasibility reference, while the second expert lacks this explicit human-event information.

\noindent \textbf{ACT.}~\cite{act}
A transformer-based action-chunking imitation-learning baseline.

\noindent \textbf{Diffusion Policy.}~\cite{diffusionpolicy}
A diffusion-based imitation-learning baseline that predicts continuous action sequences from visual observations.

\noindent\textbf{$\pi_0$}.~\cite{pi0}
An end-to-end VLA baseline pre-trained with a massive amount of robotic data.

\noindent \textbf{SmolVLA.}~\cite{smolvla}
A compact vision-language-action model designed for efficient robot policy learning, generating robot actions from multimodal observations and language instructions.

\subsection{Main Results}

\definecolor{learnedbest}{RGB}{232,239,242}
\newcommand{\best}[1]{\cellcolor{learnedbest}\textbf{#1}}
\begin{table*}[htbp]
    \centering
    \small
    \setlength{\tabcolsep}{4pt}
    \caption{\textbf{Task-level policy evaluation on \PAPERNAME.}
    We report collision-free success for each of the 37 tasks and averages over
    each condition and all tasks.
    Best nonzero values among the four learned policies in each row are highlighted,
    including ties.}
    \label{tab:main_results_vertical}
    \begin{tabular}{lcccccc}
        \toprule
        Task & \shortstack{Full-state\\expert} & \shortstack{No-intent\\expert}
             & ACT & DP & $\pi_0$ & SmolVLA \\
        \midrule
        \multicolumn{7}{l}{\textit{Cueing}} \\
        Deliver to Human & 1.00 & 0.30 & 0.00 & 0.00 & 0.00 & 0.00 \\
        Pour Water & 0.99 & 0.91 & 0.59 & \best{0.70} & 0.63 & 0.27 \\
        Oil Bottle Recovery & 0.95 & 0.01 & 0.00 & \best{0.09} & 0.07 & \best{0.09} \\
        Open Microwave & 1.00 & 1.00 & 0.20 & 0.37 & 0.39 & \best{0.45} \\
        Feed Documents & 1.00 & 0.14 & 0.00 & 0.00 & 0.00 & 0.00 \\
        Take from Human (Easy) & 0.99 & 0.32 & 0.00 & \best{0.02} & \best{0.02} & 0.00 \\
        Take from Human (Safety) & 1.00 & 0.00 & 0.00 & 0.00 & 0.00 & 0.00 \\
        Avg. & 0.99 & 0.38 & 0.11 & \best{0.17} & 0.16 & 0.11 \\
        \midrule
        \multicolumn{7}{l}{\textit{Contribution}} \\
        Categorize Objects & 1.00 & 0.47 & 0.00 & 0.00 & 0.00 & 0.00 \\
        Put Object in Cabinet & 0.95 & 0.91 & 0.06 & 0.11 & \best{0.44} & 0.05 \\
        Stack Three Bowls & 1.00 & 0.34 & 0.01 & \best{0.07} & 0.02 & 0.00 \\
        Place Bread in Basket & 1.00 & 0.53 & 0.00 & \best{0.01} & \best{0.01} & 0.00 \\
        Dump Bin & 0.98 & 0.41 & 0.05 & 0.02 & \best{0.13} & 0.00 \\
        Place Burger and Fries & 1.00 & 0.49 & 0.00 & 0.12 & \best{0.37} & 0.01 \\
        Place Pair of Shoes & 0.99 & 0.43 & 0.08 & 0.06 & \best{0.15} & 0.03 \\
        Place Food in Skillet & 0.95 & 0.20 & 0.12 & 0.02 & \best{0.30} & 0.02 \\
        Rank Blocks by Color & 1.00 & 0.86 & 0.01 & 0.00 & \best{0.10} & 0.02 \\
        Rank Blocks by Size & 1.00 & 0.73 & 0.05 & 0.00 & \best{0.12} & 0.00 \\
        Avg. & 0.99 & 0.54 & 0.04 & 0.04 & \best{0.16} & 0.01 \\
        \midrule
        \multicolumn{7}{l}{\textit{Interruption}} \\
        Categorize Objects & 1.00 & 0.00 & 0.01 & \best{0.02} & 0.00 & 0.00 \\
        Put Object in Cabinet & 1.00 & 0.00 & 0.03 & 0.00 & \best{0.11} & 0.02 \\
        Stack Three Bowls & 1.00 & 0.00 & 0.02 & 0.06 & \best{0.32} & 0.07 \\
        Place Bread in Basket & 1.00 & 0.01 & \best{0.15} & 0.14 & 0.09 & 0.00 \\
        Dump Bin & 0.98 & 0.00 & 0.01 & 0.04 & \best{0.16} & 0.00 \\
        Place Burger and Fries & 1.00 & 0.00 & 0.00 & 0.01 & \best{0.03} & 0.01 \\
        Place Pair of Shoes & 1.00 & 0.00 & 0.15 & 0.01 & \best{0.16} & 0.12 \\
        Place Food in Skillet & 1.00 & 0.00 & \best{0.51} & 0.29 & 0.06 & 0.07 \\
        Rank Blocks by Color & 1.00 & 0.02 & 0.00 & 0.00 & 0.00 & 0.00 \\
        Rank Blocks by Size & 1.00 & 0.00 & 0.00 & 0.00 & \best{0.01} & 0.00 \\
        Avg. & 1.00 & 0.00 & 0.09 & 0.06 & \best{0.09} & 0.03 \\
        \midrule
        \multicolumn{7}{l}{\textit{Co-presence}} \\
        Categorize Objects & 0.99 & 1.00 & 0.23 & \best{0.26} & 0.15 & 0.05 \\
        Put Object in Cabinet & 0.97 & 0.93 & 0.09 & \best{0.10} & 0.08 & 0.04 \\
        Stack Three Bowls & 1.00 & 1.00 & 0.77 & 0.20 & \best{0.93} & 0.68 \\
        Place Bread in Basket & 1.00 & 1.00 & 0.04 & 0.11 & \best{0.37} & 0.00 \\
        Dump Bin & 0.87 & 0.87 & 0.03 & 0.11 & \best{0.21} & 0.00 \\
        Place Burger and Fries & 1.00 & 1.00 & 0.00 & 0.14 & \best{0.45} & 0.05 \\
        Place Pair of Shoes & 1.00 & 1.00 & 0.10 & 0.05 & \best{0.27} & 0.00 \\
        Place Food in Skillet & 1.00 & 1.00 & 0.49 & 0.34 & \best{0.56} & 0.07 \\
        Rank Blocks by Color & 1.00 & 0.97 & 0.00 & 0.00 & \best{0.08} & 0.00 \\
        Rank Blocks by Size & 1.00 & 0.95 & \best{0.05} & 0.00 & \best{0.05} & 0.00 \\
        Avg. & 0.98 & 0.97 & 0.18 & 0.13 & \best{0.31} & 0.09 \\
        \midrule
        All Avg. & 0.99 & 0.48 & 0.10 & 0.09 & \best{0.19} & 0.06 \\
        \bottomrule
    \end{tabular}
\end{table*}

Table~\ref{tab:main_results_vertical} reports task success across the four human-present conditions. The full-state expert achieves 0.99 success, compared with 0.48 for the no-intent expert, indicating that the tasks are executable under the benchmark protocol. Learned policies show substantially lower success, with $\pi_0$ performing best overall at \textbf{0.19}, followed by ACT at 0.10, DP at 0.09, and SmolVLA at 0.06, with performance varying substantially across conditions. In particular, $\pi_0$ achieves 0.21 success on Dump Bin under Co-presence, but drops to 0.16 under Interruption, suggesting that human-present settings introduce policy challenges beyond standard object manipulation.

\begin{table}[!htb]
    \centering
    \small
    \setlength{\tabcolsep}{6pt}
    \caption{\textbf{Task success and safety of policies.}
    Results cover the 37 benchmark tasks.
    Success requires task completion without robot-avatar collision.
    Violat. is the mean fraction of timesteps with robot-avatar distance below
    0.10\,m; Min.\ Dist.\ is the mean episode minimum distance in meters;
    Collision is the fraction of episodes with robot-avatar contact.
    Bold marks the best value among the four learned policies in each column.}
    \label{tab:policy_diagnostics}
    \begin{tabular}{lcccc}
        \toprule
        Policy & Succ.$\uparrow$ & Violat.$\downarrow$
               & Min.\ Dist.\ (m)$\uparrow$ & Collision$\downarrow$ \\
        \midrule
        Full-state expert & 0.99 & 0.057 & 0.158 & 0.000 \\
        No-intent expert & 0.48 & 0.094 & 0.126 & 0.299 \\
        \midrule
        ACT & 0.10 & \textbf{0.080} & \textbf{0.133} & \textbf{0.102} \\
        DP & 0.09 & 0.095 & 0.120 & 0.174 \\
        $\pi_0$ & \textbf{0.19} & 0.088 & 0.095 & 0.207 \\
        SmolVLA & 0.06 & 0.093 & 0.104 & 0.214 \\
        \bottomrule
    \end{tabular}
\end{table}

\noindent \textbf{Task success and safety.}
Table~\ref{tab:policy_diagnostics} shows that higher success does not consistently improve all safety metrics. Among the four learned policies, $\pi_0$ achieves the highest success (0.19), while ACT performs best on all three safety metrics, with the lowest proximity-violation rate (0.080), the largest mean minimum separation (0.133\,m), and the lowest collision rate (0.102). Thus, success alone is insufficient to rank behavior around humans, even though collision already counts as a task failure.

\begin{table}[!t]
    \centering
    \small
    \setlength{\tabcolsep}{4pt}
    \caption{\textbf{Effect of human-present data with per-task ACT training.}
    Unlike Table~\ref{tab:main_results_vertical}, ACT is trained separately for
    each task, using 300 demonstrations per task (7,500 total per setting).
    Results are averaged within each human-present condition; Human Avg.
    summarizes these conditions.
    Parentheses show absolute gains over Base-only training.}
    \label{tab:act_single_task_variant_ablation}
    \begin{tabular}{lccccc}
        \toprule
        Training data & Base & Contribution & Interruption & Co-presence & Human Avg. \\
        \midrule
        Base-only & 0.42 & 0.10 & 0.05 & 0.14 & 0.10 \\
        Full-task & 0.41 {\color{gray}($-0.01$)}
                  & 0.18 {\color{gray}($+0.08$)}
                  & 0.08 {\color{gray}($+0.03$)}
                  & 0.19 {\color{gray}($+0.05$)}
                  & 0.15 {\color{gray}($+0.05$)} \\
        \bottomrule
    \end{tabular}
\end{table}
\noindent \textbf{Effect of Human-Present Variants.}
To isolate the effect of human activity, we compare ACT trained separately for each task on base-task data or matched human-present variant data. Each setting uses 7,500 demonstrations in total, with 300 per task across a 25-task subset; Base-only uses the paired human-absent tasks. Evaluation is performed on task families that share the same object-level goal but differ in human activity: Contribution, Interruption, and Co-presence. Table~\ref{tab:act_single_task_variant_ablation} shows that matched human-present data improves average performance on human-present tasks from 0.10 to 0.15, while base-task performance remains similar. This suggests that \PAPERNAME demonstrations help policies handle interaction-induced changes beyond the base manipulation skill.

\begin{figure}[htbp]
    \centering
    \includegraphics[width=\linewidth]{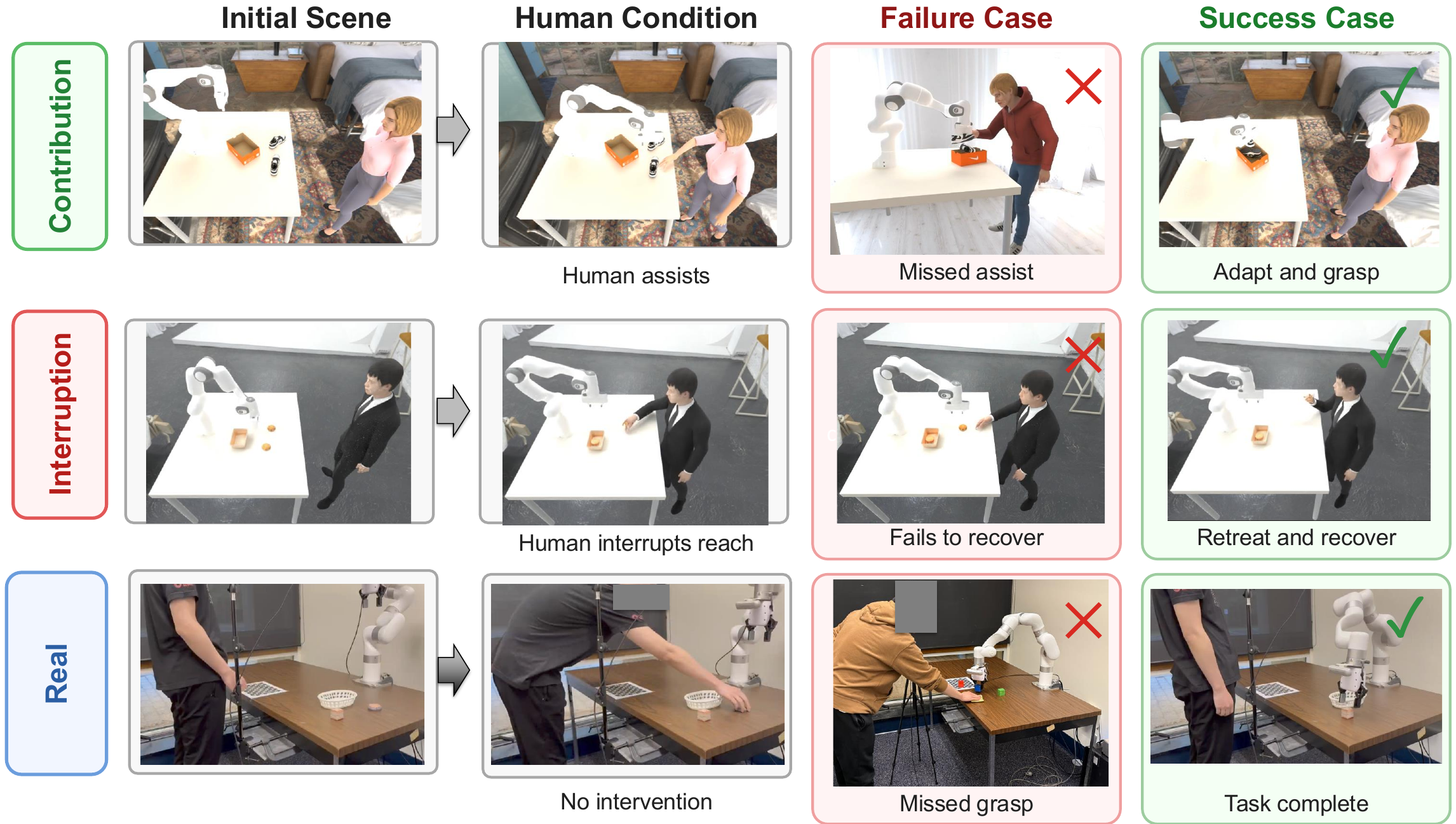}
    \caption{
    \textbf{Case studies in simulation and real-world experiments.}
    Representative rollouts illustrate missed human contributions, recovery failures after human intervention, and successful execution with human-present training data.
    }
    \label{fig:case-study}
\end{figure}

\noindent \textbf{Case Study: Failure Analysis.}
We analyze representative failures caused by human activity. As shown in Fig.~\ref{fig:case-study}, policies may fail to complete the remaining subtasks after human contribution, fail to adapt when a human changes an object's state, or violate safety around the human. In the primary failure analysis, \emph{Manipulation/recovery} includes failed manipulation and recovery after human-induced object or task-state changes; \emph{Collision} denotes robot-human collision; and \emph{Intent} denotes explicit grounding errors, such as selecting the wrong object or ignoring a cue. Manipulation/recovery accounts for 60--66\% of learned-policy failures, followed by collision at 22--27\%. $\pi_0$ retains a broadly similar failure composition despite its higher overall success rate. This does not establish identical underlying mechanisms, and the recovery category includes interaction-induced errors rather than only basic manipulation.

\subsection{Real-World Experiments}
\begin{table}[htbp]
    \centering
    \small
    \caption{
    \textbf{Real-world evaluation on the xArm dump-bin task family.} ACT success over 30 trials per evaluated setting, with matched training data sizes and steps.
    }
    \label{tab:real_world}
    \begin{tabular}{lccc}
        \toprule
        Training data & Base & Contribution & Interruption \\
        \midrule
        Base only
        & 0.40
        & 0.13
        & 0.00 \\

        Matched human-present
        & --
        & \textbf{0.33}
        & 0.00 \\
        \bottomrule
    \end{tabular}
\end{table}

\noindent \textbf{Setup.}
We evaluate sim-to-real transfer on a real xArm tabletop setup. We use the dump bin task family, which includes a base task and two human-present variants: \emph{Contribution}, where the human completes a complementary part of the task, and \emph{Interruption}, where the human inspects or occupies a task object. All real-world policies use the ACT architecture and are trained from data collected in \PAPERNAME. We selected these variants because their primary outcomes can be evaluated by task success in the same tabletop setup. Co-presence depends more directly on human-robot separation, which this setup does not measure precisely. The study tests whether \PAPERNAME data improves policy learning; it does not fully validate all four benchmark conditions or real-world safety.

\noindent \textbf{Protocol.}
We compare five train-test settings: base policy on the base task, base policy on the contribution task, base policy on the interruption task, contribution policy on the contribution task, and interruption policy on the interruption task. The base policy is trained only on robot-only demonstrations, while the contribution and interruption policies are trained on the corresponding human-present demonstrations from \PAPERNAME. Training data size and training steps are matched across settings. Each setting uses 30 trials with different object positions. For each real-world trial, the human follows the same high-level event specification as in simulation, but the exact timing and motion are not simulator-controlled.

\noindent \textbf{Results.}
Table~\ref{tab:real_world} reports real-world task success on the xArm setup. The base ACT policy achieves 0.40 success on the robot-only task, but drops to 0.13 under Contribution and 0.00 under Interruption. This shows that human-present variants also degrade policy execution in the real world. Training with human-present demonstrations from \PAPERNAME improves the Contribution variant from 0.13 to 0.33, indicating that benchmark data can help policies handle cases where the human completes part of the task. However, the Interruption policy still fails to complete the real-world interruption task, suggesting that object contention and timing around human motion remain challenging under sim-to-real transfer. Overall, these results support the need for human-present training data while also highlighting the difficulties for current policies.
\FloatBarrier
\section{Conclusion}
We presented \PAPERNAME, a benchmark of 37 human-present task instances spanning Cueing, Contribution, Interruption, and Co-presence. Its closed-loop evaluation and demonstration pipeline support both policy assessment and training. Current policies struggle with these settings: higher success does not consistently improve safety, and manipulation/recovery remains the dominant failure category. Human-present training helps in simulation and in real-world Contribution, while the negative Interruption result identifies an unresolved challenge.

\section{Limitations}
Scripted avatar events enable repeatable evaluation but do not capture continuous human-robot adaptation or the full variability of human intent, motion, and communication. The suite emphasizes tabletop manipulation, with limited interaction diversity and no systematic progression of human-interaction difficulty. It includes articulated-object interactions but does not yet cover mobile manipulation or long-horizon collaboration; the present focus is to isolate the additional challenges introduced by active humans. Sensing and contact dynamics are simplified. Success and geometric safety metrics do not directly measure appropriate waiting, rerouting, recognition of human-completed subtasks, or social comfort. Simulated safety is not evidence of safe deployment. Broader tasks, adaptive human behavior, behavior-specific metrics, and larger real-world studies remain future work.

\acknowledgments{This work was supported by MURI grant N000142412748.}


\clearpage
\appendix
\setcounter{table}{0}
\setcounter{figure}{0}
\renewcommand{\thetable}{A\arabic{table}}
\renewcommand{\thefigure}{A\arabic{figure}}
\renewcommand{\theHtable}{appendix.\arabic{table}}
\renewcommand{\theHfigure}{appendix.\arabic{figure}}
\section{Speed Benchmarking}

We benchmark avatar-rendering overhead in HARP-Bench by comparing scenes without avatars, with static avatars, and with animated avatars. Table~\ref{tab:avatar_rendering_overhead} reports rendering throughput and peak GPU memory using the Nyx backend at $640{\times}480$ resolution. All measurements are averaged over \textbf{10} episodes using the same task set on \textbf{1$\times$ NVIDIA A16 GPU}.

\subsection{Avatar Rendering Overhead}

\begin{center}
  \small
  \setlength{\tabcolsep}{4pt}
  \renewcommand{\arraystretch}{1.08}
  \begin{tabular}{lcccc}
  \toprule
  \textbf{Scene Setting} & \textbf{Backend} & \textbf{Resolution} & \textbf{FPS} & \textbf{Peak Mem.} \\
  \midrule
  No avatar & Nyx & $640{\times}480$ & 203.01 & 1.83 GB \\
  Static avatar & Nyx & $640{\times}480$ & 98.95 & 2.09 GB \\
  Animated avatar & Nyx & $640{\times}480$ & 89.55 & 2.07 GB \\
  \bottomrule
  \end{tabular}

  \vspace{2mm}
  \parbox{0.88\linewidth}{
  \small
  \captionof{table}{Avatar rendering overhead in HARP-Bench using the Nyx backend on an NVIDIA A16 GPU. We compare scenes without avatars, with static avatars, and with animated avatars. This measures the additional rendering and simulation cost introduced by human appearance and avatar motion.}
  \label{tab:avatar_rendering_overhead}
  }
\end{center}

  \paragraph{Discussion.}
  Avatar rendering introduces a clear but manageable runtime overhead. Adding a static avatar reduces rendering throughput from 203.01 FPS to 98.95 FPS, a 51.3\% decrease. Enabling avatar
  animation further reduces throughput to 89.55 FPS, corresponding to a 55.9\% decrease relative to the no-avatar scene and a 9.5\% decrease relative to the static-avatar scene. This
  indicates that most of the overhead comes from rendering the human geometry and appearance, while skeletal animation adds a smaller incremental cost. Peak GPU memory increases only
  moderately, from 1.83 GB without an avatar to about 2.1 GB with an avatar. Since HARP-Bench rollouts are recorded at 20 Hz, even the animated-avatar setting runs at roughly 4.5$\times$
  real time in this rendering benchmark, suggesting that avatar-present rollout generation remains practical. Across tasks, typical rollouts contain roughly 80--600 frames, with most tasks
  falling in the 100--300 frame range and longer interrupt or multi-stage tasks reaching about 450--600 frames, corresponding to approximately 4--30 seconds per rollout at 20 Hz.

\section{Detailed Task Definitions}

{\small
\setlength{\tabcolsep}{5pt}
\renewcommand{\arraystretch}{1.08}

\begin{longtable}{p{0.22\linewidth} p{0.70\linewidth}}
\caption{Descriptions of all 37 HARP-Bench tasks across Cueing, Contribution, Interruption, and Co-presence settings.}
\label{tab:task_descriptions}\\
\toprule
\textbf{Task} & \textbf{Description} \\
\midrule
\endfirsthead

\toprule
\textbf{Task} & \textbf{Description} \\
\midrule
\endhead

\bottomrule
\endfoot

\multicolumn{2}{l}{\textit{Cueing}} \\
\midrule

\textit{Deliver to Human} &
A requested object is placed on the table. The robot must pick up the object and deliver it to the human's hand. \\

\textit{Pour Water} &
A human drinks from a cup, cueing the robot to grasp a mug and pour water into the cup. \\

\textit{Oil Bottle Recovery} &
An oil bottle is knocked over near the cooking area. The robot must recover the bottle, place it upright, and move it to a safe region. \\

\textit{Open Microwave} &
The human carries a plate near a microwave. The robot must grasp the microwave handle, open the door, and leave the loading path clear. \\

\textit{Feed Documents} &
The human stamps two documents in sequence while the robot clears the first stamped document and pushes the second into the stamping zone. \\

\textit{Take from Human (Easy)} &
The robot retrieves a compact object presented by the human and places it on the target plate on the table. \\

\textit{Take from Human (Safety)} &
After chopping vegetables, the human offers a kitchen knife to the robot, which takes it and places it in the knife box on the robot's side of the counter. \\

\midrule
\multicolumn{2}{l}{\textit{Contribution}} \\
\midrule

\textit{Categorize Objects} &
Objects and matching baskets are placed on the table. The human and robot cooperatively sort objects into the correct baskets. \\

\textit{Put Object in Cabinet} &
The human opens the cabinet drawer, and the robot must pick up the table object and place it inside the human-opened drawer. \\

\textit{Stack Three Bowls} &
Three bowls are placed on the table. The human stacks one bowl while the robot stacks another, completing the target bowl stack. \\

\textit{Place Bread in Basket} &
Two pieces of bread and a basket are placed in the workspace. The human places one bread into the basket, and the robot places the other. \\

\textit{Dump Bin} &
The robot places table objects sampled from a mixed pool (stapler, jam jar, cube, apple, wooden block, seal, and Rubik's cube) into the bin while the human tosses in an additional object. \\

\textit{Place Burger and Fries} &
Meal items and a tray are placed on the table. The human adds an extra food item to the tray while the robot places the remaining burger and side items. \\

\textit{Place Pair of Shoes} &
A pair of shoes and a shoebox are placed on the table. The human places one shoe into the box, and the robot places the other. \\

\textit{Place Food in Skillet} &
The robot places the skillet on the stove, while the human places one food item into it. The robot must place the remaining food items into the skillet. \\

\textit{Rank Blocks by Color} &
The human places one colored block while the robot places the other two, arranging the red, green, and blue blocks in left-to-right order. \\

\textit{Rank Blocks by Size} &
The human places one block while the robot places the other two, arranging all three blocks from largest to smallest. \\

\midrule
\multicolumn{2}{l}{\textit{Interruption}} \\
\midrule

\textit{Categorize Objects} &
The robot sorts objects into matching baskets, but the human interrupts by inspecting an object the robot is about to grasp. The robot must wait, then resume sorting after the object is returned. \\

\textit{Put Object in Cabinet} &
The robot opens the cabinet drawer and prepares to place an object inside. The human inspects the object mid-task, and the robot must resume after the object is released. \\

\textit{Stack Three Bowls} &
The robot stacks bowls, but the human inspects one of the bowls during the stacking sequence. The robot must retreat, wait, and continue from the bowl's updated pose. \\

\textit{Place Bread in Basket} &
The robot places bread into a basket while the human interrupts by inspecting one bread. The robot must wait for the bread to be returned and then complete the placement. \\

\textit{Dump Bin} &
Table objects are sampled from the same mixed pool, with a trash bin beside the table. The human picks up the robot's target object; the robot must retract, adapt its timing or trajectory, and complete the bin-related goal. \\

\textit{Place Burger and Fries} &
The robot places meal items on a tray while the human inspects one food item. The robot must place the other items, wait for the inspected item, and then finish the tray arrangement. \\

\textit{Place Pair of Shoes} &
The robot places shoes into a shoebox while the human inspects one shoe mid-task. The robot must avoid the human, then resume using the shoe's updated pose. \\

\textit{Place Food in Skillet} &
The robot places a skillet and food items while the human inspects one food item. The robot must wait and then place the food into the skillet after it is returned. \\

\textit{Rank Blocks by Color} &
The robot arranges red, green, and blue blocks in left-to-right order while the human interrupts by inspecting one block. The robot must wait for the block to be returned, then complete the arrangement. \\

\textit{Rank Blocks by Size} &
The robot arranges three blocks from largest to smallest while the human inspects one block. The robot must wait for the block to be returned, then complete the arrangement. \\

\midrule
\multicolumn{2}{l}{\textit{Co-presence}} \\
\midrule

\textit{Categorize Objects} &
The robot sorts objects into matching baskets while the human performs unrelated table work nearby. \\

\textit{Put Object in Cabinet} &
The robot opens the cabinet drawer and places the object inside while the human performs unrelated table work nearby. \\

\textit{Stack Three Bowls} &
The robot stacks bowls while the human performs unrelated nearby table activity. \\

\textit{Place Bread in Basket} &
The robot places bread into the basket while the human performs unrelated table work nearby. \\

\textit{Dump Bin} &
The robot places table objects sampled from the same mixed pool into the bin while the human performs unrelated table work nearby. \\

\textit{Place Burger and Fries} &
The robot places meal items on the tray while the human performs unrelated nearby activity. \\

\textit{Place Pair of Shoes} &
The robot places both shoes into the shoebox while the human performs unrelated table work nearby. \\

\textit{Place Food in Skillet} &
The robot places the skillet and food items while the human performs unrelated table work nearby. \\

\textit{Rank Blocks by Color} &
The robot arranges red, green, and blue blocks in left-to-right order while the human performs unrelated table work nearby. \\

\textit{Rank Blocks by Size} &
The robot arranges three blocks from largest to smallest while the human performs unrelated table work nearby. \\

\end{longtable}
}

\section{\PAPERNAME Benchmark Statistics}

\subsection{Assets and Tasks}

\begin{figure}[t]
    \centering
    \includegraphics[width=\linewidth]{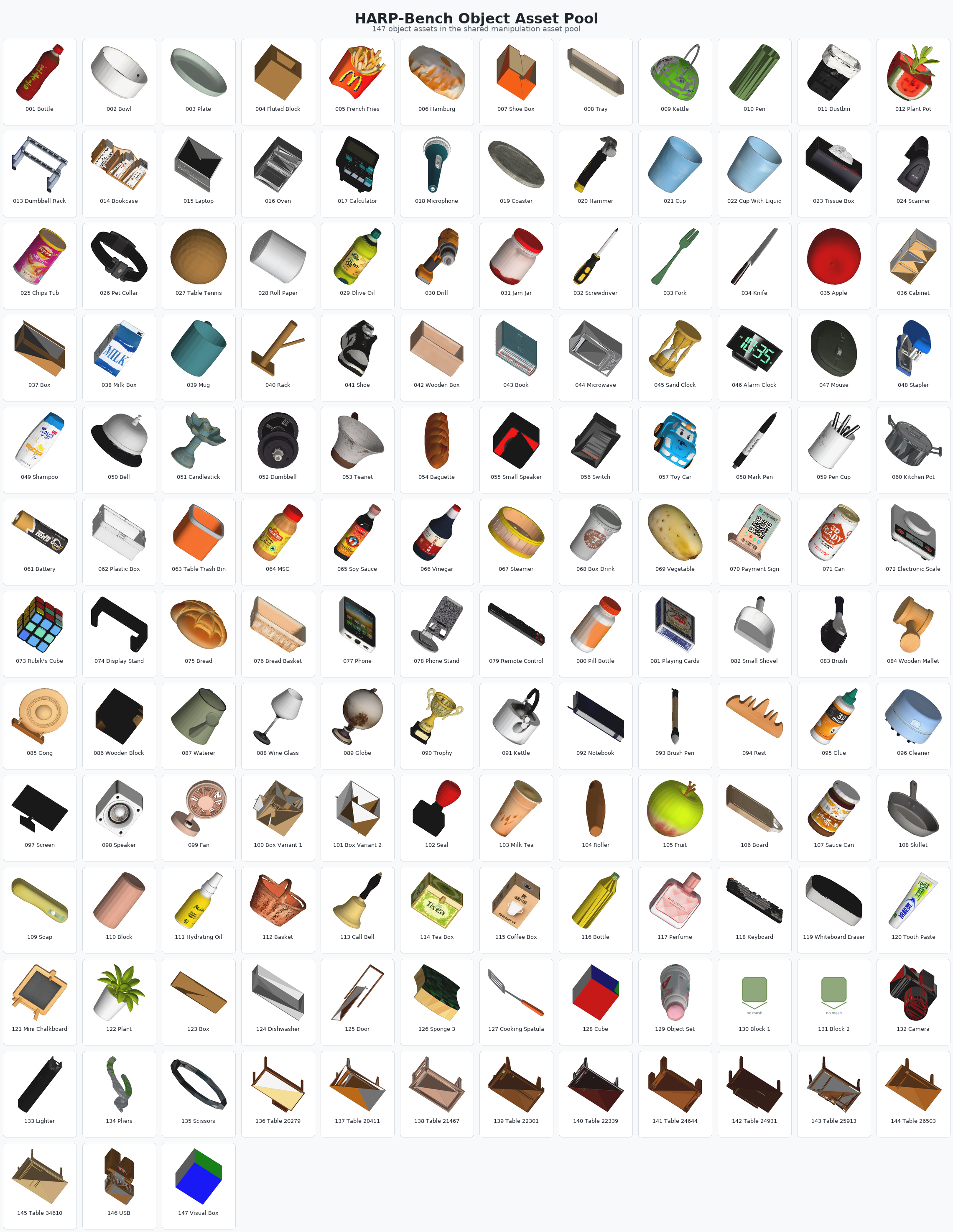}
    \caption{Object asset pool used in HARP-Bench. The pool contains 147 object asset directories, including rigid tabletop objects, articulated fixtures, containers, food items, tools, appliances, blocks, and table variants. Some assets are adapted from existing simulation asset sources, including RoboTwin-OD~\cite{robotwin2} and SAPIEN~\cite{Xiang_2020_SAPIEN, Mo_2019_CVPR, chang2015shapenet}, while others are curated or organized for HARP-Bench. Object identity is used to define task semantics and provide visual and physical variation, but not every asset appears in every rollout.}
    \label{fig:object_asset_pool}
\end{figure}

\textbf{Object asset pool.} Fig.~\ref{fig:object_asset_pool} shows the 147 object asset directories in the HARP-Bench asset pool. These assets include rigid tabletop objects, articulated fixtures, containers, food items, tools, appliances, blocks, and table variants. Each asset can be loaded by the task code as a visual and physical object in the Genesis scene, either through a mesh asset or an articulated URDF model. Task definitions specify the functional role of an object, including the target object for the robot, the target object for the human, and the container to place into. Depending on the task configuration, a rollout can use a fixed task object or sample from a task-specific subset of the pool.

\textbf{Task split.} The benchmark contains 37 human-present manipulation tasks. These tasks are grouped by the way human activity changes the robot's manipulation problem: Cueing changes what the robot should infer or do, Contribution changes which subtasks remain for the robot, Interruption changes object availability or timing, and Co-presence adds a nearby human without changing the original robot goal. The 37 tasks consist of 7 Cueing tasks, 10 Contribution tasks, 10 Interruption tasks, and 10 Co-presence tasks.

\begin{center}
\small
\setlength{\tabcolsep}{8pt}
\renewcommand{\arraystretch}{1.08}
\begin{tabular}{lc}
    \toprule
    Human-present condition & Number of tasks \\
    \midrule
    Cueing & 7 \\
    Contribution & 10 \\
    Interruption & 10 \\
    Co-presence & 10 \\
    \midrule
    Total & 37 \\
    \bottomrule
\end{tabular}

\vspace{1mm}
\captionof{table}{Task counts in the 37-task HARP-Bench benchmark.}
\label{tab:task_count_summary}
\end{center}

\textbf{Asset usage across tasks.} The 37 tasks use a shared object pool constructed from assets in SAPIEN~\cite{Xiang_2020_SAPIEN, Mo_2019_CVPR, chang2015shapenet}, RoboTwin 2.0~\cite{robotwin2}, and Sketchfab~\cite{sketchfab}. Across different tasks, the robot and human avatar interact with objects sampled from the same pool, ensuring that human-present variants are built on a consistent set of physical assets. Since imported assets can have scale mismatches when loaded into Genesis, we manually resized each object to maintain reasonable physical dimensions and task-specific usability.

\clearpage

\subsection{Avatar and Motion Assets}
\label{app:avatar_motion_assets}

\PAPERNAME uses animated avatars to generate the human activity in its human-present task variants. In the implementation, avatar appearance, avatar motion, and task-level event logic are handled separately. The avatar model determines how the human looks, the motion or control primitive determines how the avatar moves, and the task code specifies when the event occurs and how it affects the scene. This separation makes it possible to reuse the same tabletop task family across different human-present conditions while keeping the robot-side goal comparable.

\begin{figure}[p]
    \centering
    \includegraphics[width=\linewidth,height=0.86\textheight,keepaspectratio]{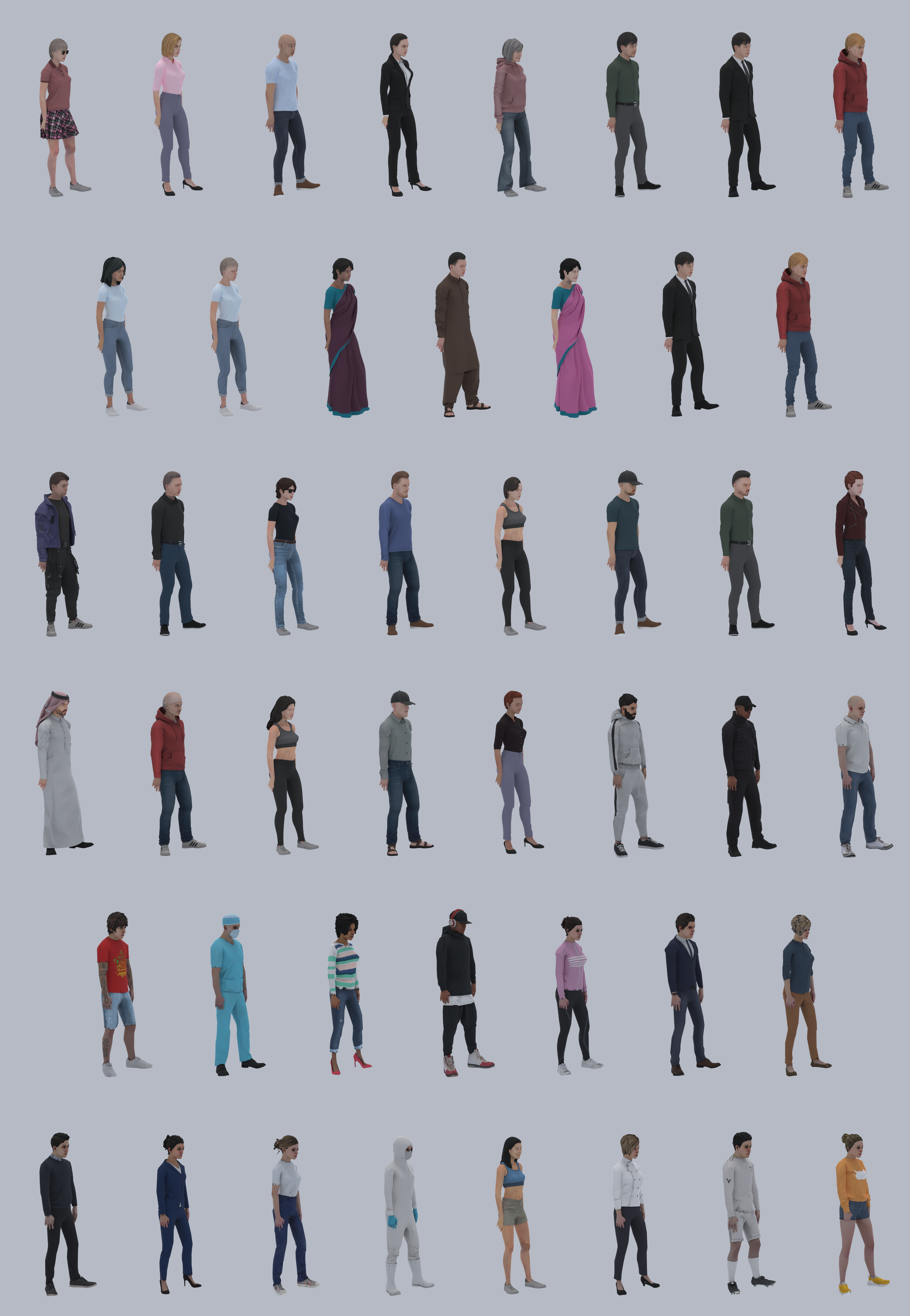}
    \caption{
    Avatar appearance pool used in HARP-Bench.
    We use 46 skinned avatar models from Virtual Community~\cite{virtualcommunity}.
    Avatar identity is used as visual appearance variation and is not tied to task semantics.
    }
    \label{fig:app_avatar_pool}
\end{figure}

\paragraph{Avatar appearance pool.}
Fig.~\ref{fig:app_avatar_pool} shows the 46 avatar appearances used in HARP-Bench. These avatar models are inherited from Virtual Community~\cite{virtualcommunity}. We adapt the original avatar simulation pipeline to work with Genesis v1.2.0~\cite{genesis2026genesisworld, Genesis}, and integrate it into our benchmark in a lightweight form, so that each task can instantiate human motion and appearance without adding unnecessary simulation overhead.

\clearpage

\paragraph{Motion and control primitives.}
Avatar behavior is built from a motion library collected from AMASS~\cite{amass} and Mixamo~\cite{mixamo}. We manually trim the downloaded motions into task-relevant clips and annotate keyframes for object interaction. At these annotated frames, the target object can be attached to or detached from the avatar hand, enabling motions such as holding, handing over, placing, tossing, or manipulating objects. In addition to replaying stored clips, we also use IK-based adjustment on top of the motion library, which allows the avatar to generate flexible pick-and-place motions conditioned on the object pose in each task instance.

\paragraph{Task-level event execution.}
The triggering logic of avatar motion depends on the task condition. In \emph{Cueing} tasks, the avatar typically performs an initial motion that indicates the intended robot action. In \emph{Contribution} tasks, the avatar starts at an early randomized time and completes part of the task before or during robot execution. In \emph{Interruption} tasks, the avatar motion is triggered based on the robot gripper position, so that the avatar randomly interferes when the robot is about to execute a relevant substep. In \emph{Co-presence} tasks, the avatar performs a randomly selected task-irrelevant motion inside the shared workspace, creating human presence without changing the robot's original goal. Fig.~\ref{fig:app_avatar_motion_primitives} summarizes the motion and control primitives used by the registered benchmark tasks.

\begin{center}
    \resizebox{!}{0.55\textheight}{%
        \vbox{\offinterlineskip
            \hbox{\includegraphics[trim=0 1423bp 0 0,clip]{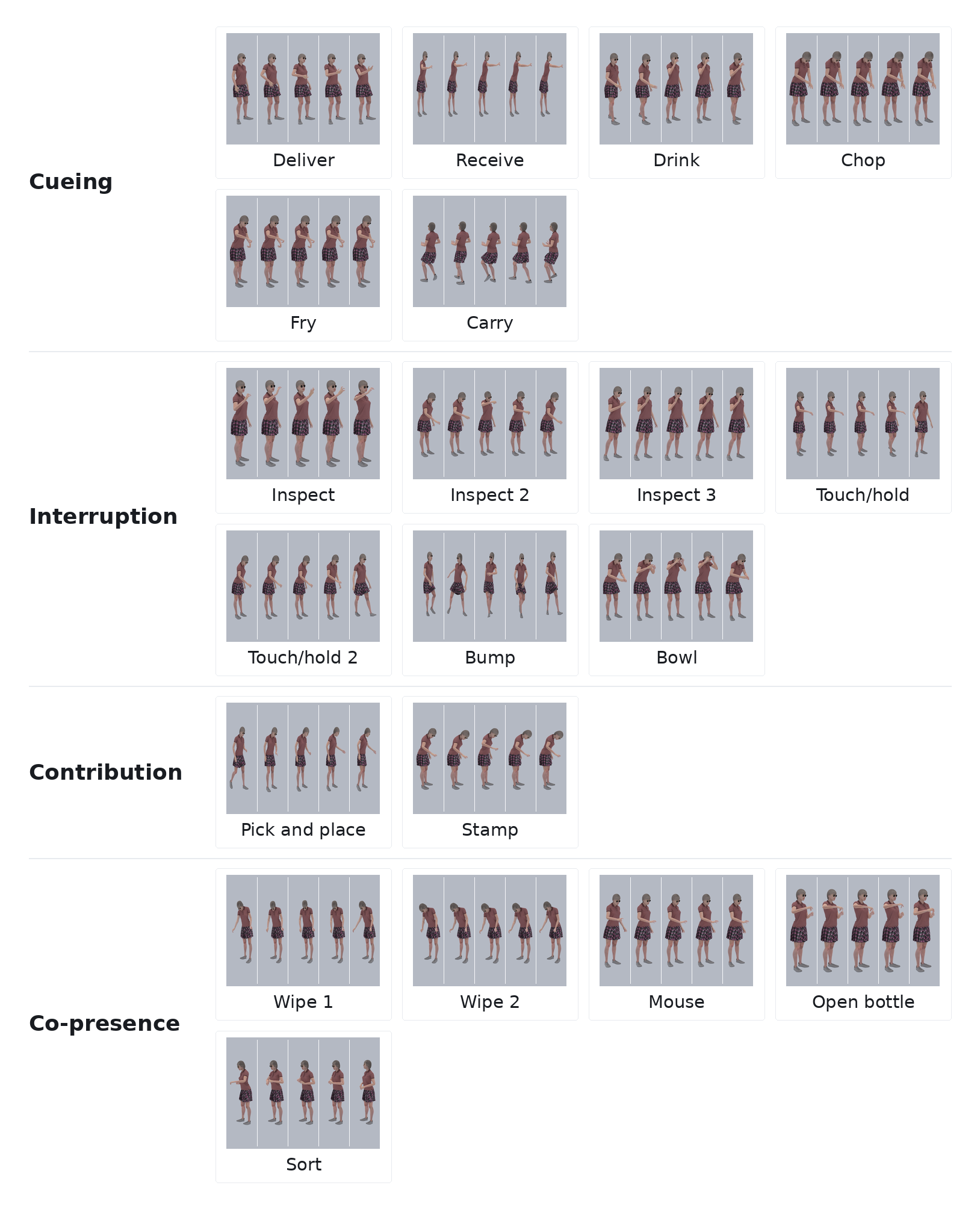}}%
            \hbox{\includegraphics[trim=0 581bp 0 1141bp,clip]{avatar_motions_all.png}}%
            \hbox{\includegraphics[trim=0 867bp 0 585bp,clip]{avatar_motions_all.png}}%
            \hbox{\includegraphics[trim=0 0 0 1427bp,clip]{avatar_motions_all.png}}%
        }%
    }

    \vspace{1mm}
    \captionof{figure}{Avatar motion and control primitives used in HARP-Bench.}
    \label{fig:app_avatar_motion_primitives}
\end{center}


\begin{thebibliography}{55}
\providecommand{\natexlab}[1]{#1}
\providecommand{\url}[1]{\texttt{#1}}
\expandafter\ifx\csname urlstyle\endcsname\relax
  \providecommand{\doi}[1]{doi: #1}\else
  \providecommand{\doi}{doi: \begingroup \urlstyle{rm}\Url}\fi

\bibitem[James et~al.(2020)James, Ma, Arrojo, and Davison]{rlbench}
S.~James, Z.~Ma, D.~R. Arrojo, and A.~J. Davison.
\newblock {RLBench}: The robot learning benchmark \& learning environment.
\newblock \emph{IEEE Robotics and Automation Letters}, 5\penalty0 (2):\penalty0
  3019--3026, 2020.

\bibitem[Zhu et~al.(2020)Zhu, Wong, Mandlekar, Mart{\'\i}n-Mart{\'\i}n, Joshi,
  Lin, Maddukuri, Nasiriany, and Zhu]{robosuite}
Y.~Zhu, J.~Wong, A.~Mandlekar, R.~Mart{\'\i}n-Mart{\'\i}n, A.~Joshi, K.~Lin,
  A.~Maddukuri, S.~Nasiriany, and Y.~Zhu.
\newblock robosuite: A modular simulation framework and benchmark for robot
  learning.
\newblock \emph{arXiv preprint arXiv:2009.12293}, 2020.

\bibitem[Mees et~al.(2022)Mees, Hermann, Rosete-Beas, and Burgard]{calvin}
O.~Mees, L.~Hermann, E.~Rosete-Beas, and W.~Burgard.
\newblock {CALVIN}: A benchmark for language-conditioned policy learning for
  long-horizon robot manipulation tasks.
\newblock \emph{IEEE Robotics and Automation Letters}, 7\penalty0 (3):\penalty0
  7327--7334, 2022.

\bibitem[Liu et~al.(2023)Liu, Zhu, Gao, Feng, Liu, Zhu, and Stone]{libero}
B.~Liu, Y.~Zhu, C.~Gao, Y.~Feng, Q.~Liu, Y.~Zhu, and P.~Stone.
\newblock {LIBERO}: Benchmarking knowledge transfer for lifelong robot
  learning.
\newblock \emph{Advances in Neural Information Processing Systems},
  36:\penalty0 44776--44791, 2023.

\bibitem[Tao et~al.(2025)Tao, Xiang, Shukla, Qin, Hinrichsen, Yuan, Bao, Lin,
  Liu, kai Chan, Gao, Li, Mu, Xiao, Gurha, Rajesh, Choi, Chen, Huang, Calandra,
  Chen, Luo, and Su]{maniskill3}
S.~Tao, F.~Xiang, A.~Shukla, Y.~Qin, X.~Hinrichsen, X.~Yuan, C.~Bao, X.~Lin,
  Y.~Liu, T.~kai Chan, Y.~Gao, X.~Li, T.~Mu, N.~Xiao, A.~Gurha, V.~N. Rajesh,
  Y.~W. Choi, Y.-R. Chen, Z.~Huang, R.~Calandra, R.~Chen, S.~Luo, and H.~Su.
\newblock {ManiSkill3}: {GPU} parallelized robotics simulation and rendering
  for generalizable embodied {AI}.
\newblock \emph{Robotics: Science and Systems}, 2025.

\bibitem[Nasiriany et~al.(2024)Nasiriany, Maddukuri, Zhang, Parikh, Lo, Joshi,
  Mandlekar, and Zhu]{robocasa}
S.~Nasiriany, A.~Maddukuri, L.~Zhang, A.~Parikh, A.~Lo, A.~Joshi, A.~Mandlekar,
  and Y.~Zhu.
\newblock {RoboCasa}: Large-scale simulation of everyday tasks for generalist
  robots.
\newblock In \emph{Robotics: Science and Systems (RSS)}, 2024.

\bibitem[Zhang et~al.(2025)Zhang, Xu, Liu, Yu, Li, Gao, Fei, Yin, Wu, Jiang,
  et~al.]{vlabench}
S.~Zhang, Z.~Xu, P.~Liu, X.~Yu, Y.~Li, Q.~Gao, Z.~Fei, Z.~Yin, Z.~Wu, Y.-G.
  Jiang, et~al.
\newblock {VLABench}: A large-scale benchmark for language-conditioned robotics
  manipulation with long-horizon reasoning tasks.
\newblock In \emph{Proceedings of the IEEE/CVF International Conference on
  Computer Vision}, pages 11142--11152, 2025.

\bibitem[Lin et~al.(2026)Lin, Zhang, Yu, Chen, Griffiths, Choi, Held, and
  Gan]{robowits}
C.~Lin, H.~Zhang, F.~Yu, Z.~Chen, T.~L. Griffiths, Y.~Choi, D.~Held, and
  C.~Gan.
\newblock {RoboWits}: Unexpected challenges for robotic creative problem
  solving.
\newblock \emph{arXiv preprint arXiv:2605.30326}, 2026.

\bibitem[Chen et~al.(2025)Chen, Chen, Chen, Cai, Liu, Li, Liang, Lin, Ge, Gu,
  et~al.]{robotwin2}
T.~Chen, Z.~Chen, B.~Chen, Z.~Cai, Y.~Liu, Z.~Li, Q.~Liang, X.~Lin, Y.~Ge,
  Z.~Gu, et~al.
\newblock {RoboTwin} 2.0: A scalable data generator and benchmark with strong
  domain randomization for robust bimanual robotic manipulation.
\newblock \emph{arXiv preprint arXiv:2506.18088}, 2025.

\bibitem[Wang et~al.(2025)Wang, Ung, Tannert, Duan, Li, Le, Oswal, Grotz,
  Pumacay, Deng, et~al.]{roboeval}
Y.~R. Wang, C.~Ung, G.~Tannert, J.~Duan, J.~Li, A.~Le, R.~Oswal, M.~Grotz,
  W.~Pumacay, Y.~Deng, et~al.
\newblock {RoboEval}: Where robotic manipulation meets structured and scalable
  evaluation.
\newblock \emph{arXiv preprint arXiv:2507.00435}, 2025.

\bibitem[Chang et~al.(2025)Chang, Chhablani, Clegg, Dallaire~Cote, Desai,
  Hlavac, Karashchuk, Krantz, Mottaghi, Parashar, et~al.]{partnr}
M.~Chang, G.~Chhablani, A.~Clegg, M.~Dallaire~Cote, R.~Desai, M.~Hlavac,
  V.~Karashchuk, J.~Krantz, R.~Mottaghi, P.~Parashar, et~al.
\newblock {PARTNR}: A benchmark for planning and reasoning in embodied
  multi-agent tasks.
\newblock In \emph{International Conference on Learning Representations},
  volume 2025, pages 65205--65268, 2025.

\bibitem[Wang et~al.(2024)Wang, Du, Xu, Li, Guo, and Liu]{humanthor}
C.~Wang, B.~Du, J.~Xu, P.~Li, D.~Guo, and H.~Liu.
\newblock Demonstrating {HumanTHOR}: a simulation platform and benchmark for
  human-robot collaboration in a shared workspace.
\newblock \emph{arXiv preprint arXiv:2406.06498}, 2024.

\bibitem[Thumm et~al.(2024)Thumm, Trost, and Althoff]{humanrobotgym}
J.~Thumm, F.~Trost, and M.~Althoff.
\newblock {Human-Robot Gym}: Benchmarking reinforcement learning in human-robot
  collaboration.
\newblock In \emph{2024 IEEE international conference on robotics and
  automation (ICRA)}, pages 7405--7411. IEEE, 2024.

\bibitem[Zhao et~al.(2023)Zhao, Kumar, Levine, and Finn]{act}
T.~Zhao, V.~Kumar, S.~Levine, and C.~Finn.
\newblock Learning fine-grained bimanual manipulation with low-cost hardware.
\newblock \emph{Robotics: Science and Systems XIX}, 2023.

\bibitem[Chi et~al.(2025)Chi, Xu, Feng, Cousineau, Du, Burchfiel, Tedrake, and
  Song]{diffusionpolicy}
C.~Chi, Z.~Xu, S.~Feng, E.~Cousineau, Y.~Du, B.~Burchfiel, R.~Tedrake, and
  S.~Song.
\newblock Diffusion policy: Visuomotor policy learning via action diffusion.
\newblock \emph{The International Journal of Robotics Research}, 44\penalty0
  (10-11):\penalty0 1684--1704, 2025.

\bibitem[Ze et~al.(2024)Ze, Zhang, Zhang, Hu, Wang, and Xu]{dp3}
Y.~Ze, G.~Zhang, K.~Zhang, C.~Hu, M.~Wang, and H.~Xu.
\newblock {3D} diffusion policy: Generalizable visuomotor policy learning via
  simple {3D} representations.
\newblock \emph{Robotics: Science and Systems XX}, 2024.

\bibitem[Zitkovich et~al.(2023)Zitkovich, Yu, Xu, Xu, Xiao, Xia, Wu, Wohlhart,
  Welker, Wahid, et~al.]{rt2}
B.~Zitkovich, T.~Yu, S.~Xu, P.~Xu, T.~Xiao, F.~Xia, J.~Wu, P.~Wohlhart,
  S.~Welker, A.~Wahid, et~al.
\newblock {RT-2}: Vision-language-action models transfer web knowledge to
  robotic control.
\newblock In \emph{Conference on Robot Learning}, pages 2165--2183. PMLR, 2023.

\bibitem[Team et~al.(2024)Team, Ghosh, Walke, Pertsch, Black, Mees, Dasari,
  Hejna, Kreiman, Xu, et~al.]{octo}
O.~M. Team, D.~Ghosh, H.~Walke, K.~Pertsch, K.~Black, O.~Mees, S.~Dasari,
  J.~Hejna, T.~Kreiman, C.~Xu, et~al.
\newblock Octo: An open-source generalist robot policy.
\newblock \emph{arXiv preprint arXiv:2405.12213}, 2024.

\bibitem[Kim et~al.(2025)Kim, Pertsch, Karamcheti, Xiao, Balakrishna, Nair,
  Rafailov, Foster, Sanketi, Vuong, et~al.]{openvla}
M.~J. Kim, K.~Pertsch, S.~Karamcheti, T.~Xiao, A.~Balakrishna, S.~Nair,
  R.~Rafailov, E.~P. Foster, P.~R. Sanketi, Q.~Vuong, et~al.
\newblock {OpenVLA}: An open-source vision-language-action model.
\newblock In \emph{Conference on Robot Learning}, pages 2679--2713. PMLR, 2025.

\bibitem[Black et~al.(2025)Black, Brown, Driess, Esmail, Equi, Finn, Fusai,
  Groom, Hausman, Ichter, et~al.]{pi0}
K.~Black, N.~Brown, D.~Driess, A.~Esmail, M.~Equi, C.~Finn, N.~Fusai, L.~Groom,
  K.~Hausman, B.~Ichter, et~al.
\newblock pi0: A vision-language-action flow model for general robot control.
\newblock \emph{Robotics: Science and Systems XXI}, 2025.

\bibitem[Intelligence et~al.(2025)Intelligence, Black, Brown, Darpinian,
  Dhabalia, Driess, Esmail, Equi, Finn, Fusai, et~al.]{pi05}
P.~Intelligence, K.~Black, N.~Brown, J.~Darpinian, K.~Dhabalia, D.~Driess,
  A.~Esmail, M.~Equi, C.~Finn, N.~Fusai, et~al.
\newblock pi0.5 a vision-language-action model with open-world generalization.
\newblock \emph{arXiv preprint arXiv:2504.16054}, 2025.

\bibitem[Liu et~al.(2025)Liu, Wu, Li, Tan, Chen, Wang, Xu, Su, and Zhu]{rdt}
S.~Liu, L.~Wu, B.~Li, H.~Tan, H.~Chen, Z.~Wang, K.~Xu, H.~Su, and J.~Zhu.
\newblock {RDT-1B}: a diffusion foundation model for bimanual manipulation.
\newblock In \emph{International Conference on Learning Representations},
  volume 2025, pages 29982--30009, 2025.

\bibitem[Lee et~al.(2025)Lee, Duan, Fang, Deng, Liu, Li, Fang, Zhang, Wang,
  Lee, et~al.]{lee2025molmoact}
J.~Lee, J.~Duan, H.~Fang, Y.~Deng, S.~Liu, B.~Li, B.~Fang, J.~Zhang, Y.~R.
  Wang, S.~Lee, et~al.
\newblock {MolmoAct}: Action reasoning models that can reason in space.
\newblock \emph{arXiv preprint arXiv:2508.07917}, 2025.

\bibitem[Bjorck et~al.(2025)Bjorck, Casta{\~n}eda, Cherniadev, Da, Ding, Fan,
  Fang, Fox, Hu, Huang, et~al.]{bjorck2025gr00t}
J.~Bjorck, F.~Casta{\~n}eda, N.~Cherniadev, X.~Da, R.~Ding, L.~Fan, Y.~Fang,
  D.~Fox, F.~Hu, S.~Huang, et~al.
\newblock {GR00T N1}: An open foundation model for generalist humanoid robots.
\newblock \emph{arXiv preprint arXiv:2503.14734}, 2025.

\bibitem[Walke et~al.(2023)Walke, Black, Lee, Kim, Du, Zheng, Zhao,
  Hansen-Estruch, Vuong, He, Myers, Fang, Finn, and Levine]{bridgedatav2}
H.~Walke, K.~Black, A.~Lee, M.~J. Kim, M.~Du, C.~Zheng, T.~Zhao,
  P.~Hansen-Estruch, Q.~Vuong, A.~He, V.~Myers, K.~Fang, C.~Finn, and
  S.~Levine.
\newblock {BridgeData V2}: A dataset for robot learning at scale.
\newblock In \emph{Conference on Robot Learning (CoRL)}, 2023.

\bibitem[Khazatsky et~al.(2024)Khazatsky, Pertsch, Nair, Balakrishna, Dasari,
  Karamcheti, Nasiriany, Srirama, Chen, Ellis, et~al.]{droid}
A.~Khazatsky, K.~Pertsch, S.~Nair, A.~Balakrishna, S.~Dasari, S.~Karamcheti,
  S.~Nasiriany, M.~K. Srirama, L.~Y. Chen, K.~Ellis, et~al.
\newblock {DROID}: A large-scale in-the-wild robot manipulation dataset.
\newblock \emph{arXiv preprint arXiv:2403.12945}, 2024.

\bibitem[O’Neill et~al.(2024)O’Neill, Rehman, Maddukuri, Gupta, Padalkar,
  Lee, Pooley, Gupta, Mandlekar, Jain, et~al.]{openxembodiment}
A.~O’Neill, A.~Rehman, A.~Maddukuri, A.~Gupta, A.~Padalkar, A.~Lee,
  A.~Pooley, A.~Gupta, A.~Mandlekar, A.~Jain, et~al.
\newblock {Open X-Embodiment}: Robotic learning datasets and {RT-X} models:
  Open x-embodiment collaboration 0.
\newblock In \emph{2024 IEEE International Conference on Robotics and
  Automation (ICRA)}, pages 6892--6903. IEEE, 2024.

\bibitem[Cadene et~al.(2026)Cadene, Aliberts, Capuano, Aractingi, Zouitine,
  Kooijmans, Choghari, Russi, Pascal, Palma, et~al.]{lerobot}
R.~Cadene, S.~Aliberts, F.~Capuano, M.~Aractingi, A.~Zouitine, P.~Kooijmans,
  J.~Choghari, M.~Russi, C.~Pascal, S.~Palma, et~al.
\newblock {LeRobot}: An open-source library for end-to-end robot learning.
\newblock \emph{arXiv preprint arXiv:2602.22818}, 2026.

\bibitem[Wang et~al.(2026)Wang, Zhang, Liu, Zhang, Cai, Liu, and Liu]{liberox}
G.~Wang, C.~Zhang, Q.~Liu, J.~Zhang, J.~Cai, J.~Liu, and X.~Liu.
\newblock {LIBERO-X}: Robustness litmus for vision-language-action models.
\newblock \emph{arXiv preprint arXiv:2602.06556}, 2026.

\bibitem[Nasiriany et~al.(2026)Nasiriany, Nasiriany, Maddukuri, and
  Zhu]{robocasa365}
S.~Nasiriany, S.~Nasiriany, A.~Maddukuri, and Y.~Zhu.
\newblock {RoboCasa365}: A large-scale simulation framework for training and
  benchmarking generalist robots.
\newblock In \emph{International Conference on Learning Representations
  (ICLR)}, 2026.

\bibitem[Zhang et~al.(2025)Zhang, Li, Shen, Cai, Zhang, Chen, Dai, Ji, and
  Yang]{vlaarena}
B.~Zhang, J.~Li, J.~Shen, Y.~Cai, Y.~Zhang, Y.~Chen, J.~Dai, J.~Ji, and
  Y.~Yang.
\newblock {VLA-Arena}: An open-source framework for benchmarking
  vision-language-action models.
\newblock \emph{arXiv preprint arXiv:2512.22539}, 2025.

\bibitem[Atreya et~al.(2025)Atreya, Pertsch, Lee, Kim, Jain, Kuramshin, Eppner,
  Neary, Hu, Ramos, et~al.]{atreya2025roboarena}
P.~Atreya, K.~Pertsch, T.~Lee, M.~J. Kim, A.~Jain, A.~Kuramshin, C.~Eppner,
  C.~Neary, E.~Hu, F.~Ramos, et~al.
\newblock {RoboArena}: Distributed real-world evaluation of generalist robot
  policies.
\newblock \emph{arXiv preprint arXiv:2506.18123}, 2025.

\bibitem[Li et~al.(2023)Li, Zhang, Wong, Gokmen, Srivastava,
  Mart{\'\i}n-Mart{\'\i}n, Wang, Levine, Lingelbach, Sun, et~al.]{behavior1k}
C.~Li, R.~Zhang, J.~Wong, C.~Gokmen, S.~Srivastava, R.~Mart{\'\i}n-Mart{\'\i}n,
  C.~Wang, G.~Levine, M.~Lingelbach, J.~Sun, et~al.
\newblock {BEHAVIOR-1K}: A benchmark for embodied {AI} with 1,000 everyday
  activities and realistic simulation.
\newblock In \emph{Conference on Robot Learning}, pages 80--93. PMLR, 2023.

\bibitem[Yoshida et~al.(2025)Yoshida, Dossa, Di~Vincenzo, Sujit, Douglas, and
  Arulkumaran]{m4bench}
A.~Yoshida, R.~F.~J. Dossa, M.~Di~Vincenzo, S.~Sujit, H.~Douglas, and
  K.~Arulkumaran.
\newblock A multi-user multi-robot multi-goal multi-device human-robot
  interaction manipulation benchmark.
\newblock \emph{Frontiers in Robotics and AI}, 12:\penalty0 1528754, 2025.

\bibitem[Zhang et~al.(2026)Zhang, Bai, Zhou, Zhang, Zhang, Luan, and
  Chen]{niabench}
Y.~Zhang, S.~Bai, W.~Zhou, H.~Zhang, Q.~Zhang, Z.~Luan, and B.~Chen.
\newblock Assistance without interruption: A benchmark and {LLM}-based
  framework for non-intrusive human-robot assistance.
\newblock \emph{arXiv preprint arXiv:2605.01368}, 2026.

\bibitem[Zhang et~al.(2025)Zhang, Dong, Bai, Ni, Marton, Chen, and
  Zhang]{responsiblerobotbench}
L.~Zhang, J.~Dong, K.~Bai, M.~Ni, Z.-C. Marton, Z.~Chen, and J.~Zhang.
\newblock {ResponsibleRobotBench}: Benchmarking responsible robot manipulation
  using multi-modal large language models.
\newblock \emph{arXiv preprint arXiv:2512.04308}, 2025.

\bibitem[Zhang et~al.(2024)Zhang, Du, Shan, Zhou, Du, Tenenbaum, Shu, and
  Gan]{zhang2024building}
H.~Zhang, W.~Du, J.~Shan, Q.~Zhou, Y.~Du, J.~B. Tenenbaum, T.~Shu, and C.~Gan.
\newblock Building cooperative embodied agents modularly with large language
  models.
\newblock In \emph{International Conference on Learning Representations},
  volume 2024, pages 19373--19401, 2024.

\bibitem[Zhang et~al.(2025)Zhang, Wang, Lyu, Zhang, Chen, Shu, Dariush, Lee,
  Du, and Gan]{zhang2025combo}
H.~Zhang, Z.~Wang, Q.~Lyu, Z.~Zhang, S.~Chen, T.~Shu, B.~Dariush, K.~Lee,
  Y.~Du, and C.~Gan.
\newblock {COMBO}: Compositional world models for embodied multi-agent
  cooperation.
\newblock In \emph{International Conference on Learning Representations},
  volume 2025, pages 49996--50019, 2025.

\bibitem[Carroll et~al.(2019)Carroll, Shah, Ho, Griffiths, Seshia, Abbeel, and
  Dragan]{carroll2019utility}
M.~Carroll, R.~Shah, M.~K. Ho, T.~Griffiths, S.~Seshia, P.~Abbeel, and
  A.~Dragan.
\newblock On the utility of learning about humans for human-{AI} coordination.
\newblock \emph{Advances in neural information processing systems}, 32, 2019.

\bibitem[Wang et~al.(2024)Wang, Kedia, Ren, Abdullah, Bhardwaj, Chao, Chen,
  Chin, Dan, Fan, et~al.]{mosaic}
H.~Wang, K.~Kedia, J.~Ren, R.~Abdullah, A.~Bhardwaj, A.~Chao, K.~Y. Chen,
  N.~Chin, P.~Dan, X.~Fan, et~al.
\newblock {MOSAIC}: A modular system for assistive and interactive cooking.
\newblock In \emph{2nd Workshop on Mobile Manipulation and Embodied
  Intelligence at ICRA 2024}, 2024.

\bibitem[De~Lazzari et~al.(2025)De~Lazzari, Terreran, Giacomuzzo, Jain, Falco,
  Carli, and Romeres]{pace}
D.~De~Lazzari, M.~Terreran, G.~Giacomuzzo, S.~Jain, P.~Falco, R.~Carli, and
  D.~Romeres.
\newblock {PACE}: proactive assistance in human-robot collaboration through
  action-completion estimation.
\newblock In \emph{2025 IEEE international conference on robotics and
  automation (ICRA)}, pages 6725--6731. IEEE, 2025.

\bibitem[Giacomuzzo et~al.(2024)Giacomuzzo, Terreran, Jain, and Romeres]{decaf}
G.~Giacomuzzo, M.~Terreran, S.~Jain, and D.~Romeres.
\newblock {DECAF}: a discrete-event based collaborative human-robot framework
  for furniture assembly.
\newblock In \emph{2024 IEEE/RSJ international conference on intelligent robots
  and systems (IROS)}, pages 7085--7091. IEEE, 2024.

\bibitem[Zhou et~al.(2026)Zhou, Zhang, Lin, Zhang, Chen, Liu, Zhang, Chen,
  Fang, Lyu, Sun, Yang, Wang, Dang, Chen, Ladia, Dang, Liu, and
  Gan]{virtualcommunity}
Q.~Zhou, H.~Zhang, X.~Lin, Z.~Zhang, Y.~Chen, W.~Liu, Z.~Zhang, S.~Chen,
  L.~Fang, Q.~Lyu, X.~Sun, J.~Yang, Z.~Wang, B.~C. Dang, Z.~Chen, D.~Ladia,
  Q.~V. Dang, J.~Liu, and C.~Gan.
\newblock {Virtual Community}: An open world for humans, robots, and society.
\newblock In \emph{The Fourteenth International Conference on Learning
  Representations}, 2026.
\newblock URL \url{https://openreview.net/forum?id=Qo0OZZoTLh}.

\bibitem[Mahmood et~al.(2019)Mahmood, Ghorbani, Troje, Pons-Moll, and
  Black]{amass}
N.~Mahmood, N.~Ghorbani, N.~F. Troje, G.~Pons-Moll, and M.~J. Black.
\newblock {AMASS}: Archive of motion capture as surface shapes.
\newblock In \emph{International Conference on Computer Vision}, pages
  5442--5451, Oct. 2019.

\bibitem[{Adobe}(2026)]{mixamo}
{Adobe}.
\newblock Mixamo.
\newblock \url{https://www.mixamo.com/}, 2026.
\newblock Accessed: 2026-05-25.

\bibitem[Hoffman et~al.(2024)Hoffman, Bhattacharjee, and
  Nikolaidis]{hoffman2024inferring}
G.~Hoffman, T.~Bhattacharjee, and S.~Nikolaidis.
\newblock Inferring human intent and predicting human action in human--robot
  collaboration.
\newblock \emph{Annual Review of Control, Robotics, and Autonomous Systems},
  7:\penalty0 73--95, 2024.
\newblock \doi{10.1146/annurev-control-071223-105834}.

\bibitem[Lasota and Shah(2015)]{lasota2015analyzing}
P.~A. Lasota and J.~A. Shah.
\newblock Analyzing the effects of human-aware motion planning on
  close-proximity human--robot collaboration.
\newblock \emph{Human Factors}, 57\penalty0 (1):\penalty0 21--33, 2015.
\newblock \doi{10.1177/0018720814565188}.

\bibitem[{International Organization for Standardization}(2016)]{iso15066}
{International Organization for Standardization}.
\newblock Robots and robotic devices---collaborative robots.
\newblock Technical Specification ISO/TS 15066:2016, International Organization
  for Standardization, Feb. 2016.
\newblock URL \url{https://www.iso.org/standard/62996.html}.

\bibitem[Shukor et~al.(2025)Shukor, Aubakirova, Capuano, Kooijmans, Palma,
  Zouitine, Aractingi, Pascal, Russi, Marafioti, et~al.]{smolvla}
M.~Shukor, D.~Aubakirova, F.~Capuano, P.~Kooijmans, S.~Palma, A.~Zouitine,
  M.~Aractingi, C.~Pascal, M.~Russi, A.~Marafioti, et~al.
\newblock {SmolVLA}: A vision-language-action model for affordable and
  efficient robotics.
\newblock \emph{arXiv preprint arXiv:2506.01844}, 2025.

\bibitem[Xiang et~al.(2020)Xiang, Qin, Mo, Xia, Zhu, Liu, Liu, Jiang, Yuan,
  Wang, Yi, Chang, Guibas, and Su]{Xiang_2020_SAPIEN}
F.~Xiang, Y.~Qin, K.~Mo, Y.~Xia, H.~Zhu, F.~Liu, M.~Liu, H.~Jiang, Y.~Yuan,
  H.~Wang, L.~Yi, A.~X. Chang, L.~J. Guibas, and H.~Su.
\newblock {SAPIEN}: A simulated part-based interactive environment.
\newblock In \emph{The IEEE Conference on Computer Vision and Pattern
  Recognition (CVPR)}, June 2020.

\bibitem[Mo et~al.(2019)Mo, Zhu, Chang, Yi, Tripathi, Guibas, and
  Su]{Mo_2019_CVPR}
K.~Mo, S.~Zhu, A.~X. Chang, L.~Yi, S.~Tripathi, L.~J. Guibas, and H.~Su.
\newblock {PartNet}: A large-scale benchmark for fine-grained and hierarchical
  part-level {3D} object understanding.
\newblock In \emph{The IEEE Conference on Computer Vision and Pattern
  Recognition (CVPR)}, June 2019.

\bibitem[Chang et~al.(2015)Chang, Funkhouser, Guibas, Hanrahan, Huang, Li,
  Savarese, Savva, Song, Su, et~al.]{chang2015shapenet}
A.~X. Chang, T.~Funkhouser, L.~Guibas, P.~Hanrahan, Q.~Huang, Z.~Li,
  S.~Savarese, M.~Savva, S.~Song, H.~Su, et~al.
\newblock {ShapeNet}: An information-rich {3D} model repository.
\newblock \emph{arXiv preprint arXiv:1512.03012}, 2015.

\bibitem[{Sketchfab}()]{sketchfab}
{Sketchfab}.
\newblock Sketchfab.
\newblock \url{https://sketchfab.com/}.

\bibitem[Team(2026)]{genesis2026genesisworld}
G.~A. Team.
\newblock The role of simulation in scalable robotics, {Genesis World} 1.0, and
  the path forward.
\newblock \emph{Genesis AI Blog}, May 2026.
\newblock URL
  \url{https://www.genesis.ai/blog/the-role-of-simulation-in-scalable-robotics-genesis-world-10-and-the-path-forward}.

\bibitem[Authors(2024)]{Genesis}
G.~Authors.
\newblock {Genesis}: A generative and universal physics engine for robotics and
  beyond, December 2024.
\newblock URL \url{https://github.com/Genesis-Embodied-AI/genesis-world}.

\end{thebibliography}
\end{document}